\documentclass[final]{elsarticle}
\makeatletter
\def\ps@pprintTitle{%
 \let\@oddhead\@empty
 \let\@evenhead\@empty
 \def\@oddfoot{\centerline{\thepage}}%
 \let\@evenfoot\@oddfoot}
\makeatother

\usepackage{graphicx}
\usepackage{amsmath}
\usepackage{amssymb}

\usepackage{booktabs}
\usepackage{verbatim}
\usepackage{subcaption}
\usepackage{makecell}
\usepackage{xcolor}
\usepackage{multirow}
\usepackage{subcaption}

\usepackage{array, caption, floatrow, tabularx, makecell, booktabs}%

\begin{document}

\begin{frontmatter}
\title{Linguistically-aware Attention for Reducing the Semantic-Gap in Vision-Language Tasks}
\author[1]{Gouthaman KV}
\author[2]{Athira Nambiar}
\author[3]{Kancheti Sai Srinivas}
\author[4]{Anurag Mittal}
\address{Department of Computer Science and Engineering\\Indian Institute of Technology Madras, Chennai, India}
 \address[1]{gkv@cse.iitm.ac.in}
 \address[2]{anambiar@cse.iitm.ac.in}
 \address[3]{coolsai.srini@gmail.com}
 \address[4]{amittal@cse.iitm.ac.in}

\begin{abstract}
Attention models are widely used in Vision-language (V-L) tasks to perform the visual-textual correlation.  Humans perform such a correlation with a strong linguistic understanding of the visual world.
However, even the best performing attention model in V-L tasks lacks such a high-level linguistic understanding, thus creating a semantic gap between the modalities.
In this paper, we propose an attention mechanism - Linguistically-aware Attention (LAT) - that leverages object attributes obtained from generic object detectors along with pre-trained language models to reduce this semantic gap.  LAT represents visual and textual modalities in a common linguistically-rich space, thus providing linguistic awareness to the attention process.   We apply and demonstrate the effectiveness of LAT in three V-L tasks: Counting-VQA, VQA, and Image captioning.  In Counting-VQA, we propose a novel counting-specific VQA model to predict an intuitive count and achieve state-of-the-art results on five datasets.  In VQA and Captioning, we show the generic nature and effectiveness of LAT by adapting it into various baselines and consistently improving their performance.   
\end{abstract}


\end{frontmatter}

\section{Introduction}
Multi-modal problems involving Computer Vision and Natural Language Processing is an important area inviting a lot of attention from the AI community.
Addressing such problems in the current technological world is quite significant wherein a human user can easily interact with a  machine such as a chat-box or a robot in a human-friendly manner and thus bridge the gap between human and machine interpretations of the world.  To achieve this goal, the models require
scene and language understanding capabilities and a joint understanding of the two.  Hence, such problems are a good metric for testing whether computers are reaching a human-level understanding of the world.

Visual Question Answering (VQA) and Image Captioning are two well known yet challenging problems in the Vision-Language (V-L) domain.   VQA is the task of answering a question from the image~\cite{vqa1}, whereas captioning needs to describe the content of an image as a short natural language sentence~\cite{karpathy2015deep}.
Further, counting questions in VQA has also been recognized as a separate problem in the community as ``Counting-VQA" and has been studied using specialized VQA models~\cite{irlc,tallyqa,iclr2} as general-purpose VQA models fail to perform well on such questions.
In this paper, we address these three V-L problems.

The most successful and widely adopted approach for the V-L problems is the encoder-decoder approach~\cite{vqa1,karpathy2015deep,bottomup}.  In such an approach, the visual and textual modalities are converted to feature vectors in the encoding stage, and the desired outputs are predicted from them in the decoding stage.   Early approaches use CNNs over the whole image to encode the image and RNNs (LSTM or GRU) over the sequence of word-level features to encode the text (question or caption).  
Then, these encoded vectors are projected to some learned common space to conduct further reasoning~\cite{What_do_explicit,wuKb,vqa1}.  One downside of this feature extraction is that the whole-image CNN features may contribute noise to the model since all the image regions may not be equally important in the context, and there may be clutter in the background that will also get encoded.

To overcome this limitation, attention mechanisms have been proposed~\cite{xu2015show} to facilitate fine-grain visual and/or textual processing.   Attention models have enabled the system to better focus on the relevant image regions~\cite{xu2015show,san,ban}, text words, or both~\cite{ban} and imparted significant improvements to the field.
Some early approaches perform the attention over equally-sized image grids~\cite{san}.  However, such methods lack the whole-object-level information during attention.  Recently, Anderson \textit{et al.}~\cite{bottomup} proposed object-level attention by considering the image as a set of object proposals obtained from pre-trained generic object detectors.   They demonstrate the transfer learning possibilities from the object-detection research to V-L tasks, and the extracted object-level CNN features are generic and easy to use.  Since then, object-level CNN features became the chosen image representation in many recent state-of-the-art V-L models for various tasks~\cite{ban,aoa,tallyqa}.

\begin{figure}[t]
    \centering
   \scalebox{0.9}{
   \includegraphics[width=0.8\columnwidth]{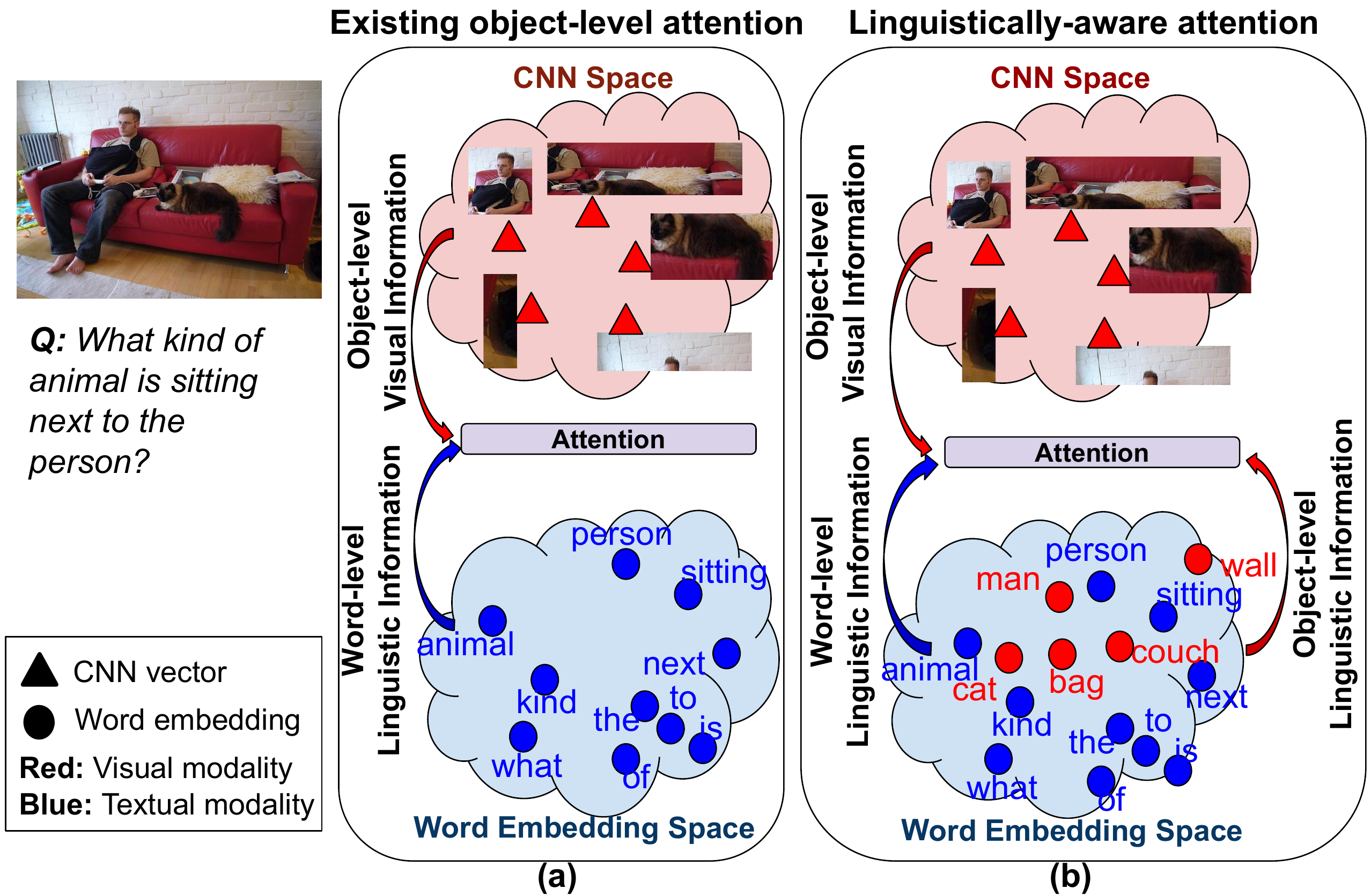}}
  \caption{Existing object-level attention \textit{vs.} Linguistically-aware attention (Ours): We provide linguistic awareness to the attention by representing both the Image (via object class labels) and Text (via words) in a common pre-trained word embedding space.   Hence, the linguistically related objects and words will get similar representations.   Best viewed in color.
  }
  \label{fig:intro}
\end{figure}

However, representing the image as a set of object-level CNN features and the text as a set of word-embedding vectors has a limitation that the two modalities are in two different spaces.  Such a semantic-gap between the modalities inhibits the performance of the V-L tasks.
Humans perform visual-textual mapping with a strong linguistic understanding of the visual data, and this suggests that V-L systems should do the same to improve performance.
For instance, consider the VQA problem in Fig~\ref{fig:intro} to answer the question:\textit{``What kind of animal is sitting next to the person?"}.  For humans, it is quite easy to interpret the linguistic relationship of the \textit{``cat"} in the image with the \textit{``animal"} in the question since we have a strong linguistic understanding of the relationship between the two words.   However, it is difficult for the existing object-level attention mechanisms to draw such a linguistic relationship since there is no explicit object-level linguistic information available for the mapping (Fig~\ref{fig:intro}(a)).  As a result, the correlation has to be done between two different spaces (visual and textual).   We conjecture that the existence of such a semantic gap between the two modalities leads to sub-optimal attention modeling.

One possible way to reduce this semantic gap is to represent the modalities in a common space, i.e., representing the text in the visual space or the Image in the textual space.   Representing the text in the visual space is a more challenging problem because of the huge dimensionality of the visual space.   An easier method is the reverse, i.e., representing the image in the textual space.   To this end, object attributes (class labels) have been used to represent the image in the textual space.
Most of the prior work on object attributes uses detectors trained specifically on a word vocabulary extracted from specific datasets of the respective tasks~\cite{What_do_explicit,mlin}.  
However, creating a dataset-specific attribute vocabulary for each task and dataset is not practically feasible, and it restricts the models from real-world application scenarios.  Due to these engineering complexities, the usage of object-attributes in V-L tasks has not gained much popularity.    

In this paper, we study the use of object attributes obtained from general-purpose object detectors such as YOLO~\cite{yolo} and Faster-RCNN~\cite{fasterrcnn} to reduce the semantic-gap in V-L problems.   Although such pre-trained object detectors are easy to use and are robust, one downside is that they can only predict the object classes that they are trained on; conversely, for some unknown class objects, they predict a similar known class only.   For instance, for a Counting-VQA problem ``How many sedans are there?", given an image with sedan cars, the object detector may predict the class ``car" if it is unaware of the class ``sedan."  Hence, the Counting-VQA model cannot relate the word ``sedan" in the question to the word ``car" in the image unless it is provided with some external knowledge-base or a training set with a lot of such samples.   To overcome this problem, we propose to leverage the same \textit{pre-trained word embedding networks} used for the question words to represent the object attributes as well.   Note that such word-embedding vectors are inherently linguistically-rich since they are trained using a large text corpus and are created in a way that semantically related words will get similar vector representations.   Hence, even if the object detector is unaware of some classes, the use of pre-trained word embeddings ensures that the known class will get a representation similar to the original class.   As a result, the word ``sedan" and image ``car" will have similar representations.  
In this way, we can utilize the recent progress in the object-detection and the language modeling research into the V-L problems to reduce the semantic gap between the modalities.

We term the pre-trained word embedding vectors of object attributes as Linguistically-aware image features and leverage them in the attention modeling: the 
\textbf{Linguistically-aware Attention (LAT)}.   A visual comparison of LAT with existing object-level attention is shown in Fig~\ref{fig:intro}.   In particular, LAT (Fig~\ref{fig:intro}(b)) represents the objects not only in the CNN space but also in the word-embedding space via the object class labels.   The words of the text are also represented in the same word embedding space.   As a result, semantically similar objects and words have similar representations (In Fig~\ref{fig:intro}(b), the ``cat" lies close to the word ``animal" in the common word-embedding space), thus making the mapping easy and linguistically robust.   Since LAT uses generic object detectors (as opposed to task-specific detectors), it is generic and easy to use for various V-L tasks.  In this paper, we show the effectiveness of LAT in three V-L tasks:
\begin{itemize}
    \item \textbf{Counting-VQA}: We propose a novel model that consists of a \textit{semantic-level co-attention} (consists of LAT) and a low-rank \textit{tensor regression-based count predictor} (for the first time to the best of our knowledge).  
    Our model achieves state-of-the-art results on five counting specific VQA datasets.
    \item \textbf{VQA}: In this task, we show the generic nature of LAT by adapting it to recent best performing VQA models: UpDn~\cite{bottomup}, MUREL~\cite{murel} and BAN~\cite{ban}.  In all the models, LAT consistently improves the accuracy.
    \item \textbf{Captioning}: For this task, we adapt LAT into the best performing object-level attention-based model~\cite{bottomup}.  Here also, LAT improves the performance.
\end{itemize}

\noindent We believe that \textit{linguistically-aware image features} can be used as standard features like the object-level CNN features for a variety of V-L tasks as our approach is simple and can be adapted to a variety of V-L tasks.  Thus, the paper is well suited for this special issue, as it focuses on methods to reduce the semantic-gap in multi-modal problems; our work is best suitable for the same.  

\section{Related works}
\noindent \textbf{Visual question answering (VQA): }
VQA is the task of answering a question related to an image~\cite{vqa1}.  
The encoder-decoder approach is the most popular scheme in which the image and the question are encoded in the encoding stage via a pre-trained CNN~\cite{vqa1} and RNN (LSTM or  GRU)~\cite{vqa1,bottomup} respectively.   The decoder further fuses them using any standard fusion strategy, such as element-wise multiplication/sum, concatenation, or bi-linear operations~\cite{mcb,mutan} and then performs a classification to one of the answers in a pre-defined set of answers.
Following~\cite{bottomup}, the usage of CNN features from the objects in the encoder has become popular, and many recent works are built upon such a framework.   For example, \cite{ban} uses bi-linear techniques to model the co-attention between the objects and the question words, 
~\cite{gcn_vqa} proposed a graph convolutional network-based approach to encode the image by incorporating the relationships among objects, \cite{murel} exploited a rich vector representation between the question and the objects to model pair-wise object relations explicitly, 
\cite{mlin} finds a summarized vector representation for each modality and aggregates it with the object and word information to obtain a better-encoded vector to predict the answer.   Some approaches, such as LXMERT~\cite{lxmert} and ViLBERT~\cite{vilbert}, use Transformer-based co-attention on top of the object-level CNN and question word features.

\noindent \textbf{Counting-VQA:}
In VQA, the questions are categorized into three:  \textit{yes/no}, \textit{number} (answer is in the form of a number) and \textit{others} (all the types other than the above two) \cite{vqa1}.  A majority of the \textit{number} category questions are of counting type, wherein the system has to enumerate the objects in the image satisfying certain criteria in the question (\textit{e.g., How many zebras are there?}).   While excellent results have been obtained for non-counting questions~\cite{bottomup}, the significant progress in the field in recent years has not reflected in the accuracy achieved for the counting questions so far.   This restricts the current VQA models from being applicable in practical scenarios.

The under-performance of general VQA models on counting questions led many researchers to look into this problem separately as Counting-VQA.   For instance, \cite{iclr2} and \cite{tallyqa} considered counting as a classification problem similar to the general VQA problem.  Specifically, \cite{iclr2} constructed a graph of objects and removed edges via several heuristics to estimate the number of objects that match the question.  Similarly, \cite{tallyqa} proposed a model that predicts the count by modeling the relationship between the objects and the background regions in the image.  \cite{everyday} came up with a dedicated generic counting model that counts the occurrences of all the object categories present in the image.  Further, they enumerated the question-specific objects by finding the closest object categories to the first noun of the question.  \cite{irlc} proposed a model that can predict an intuitive count as the answer by extending the VQA model from~\cite{bottomup}.

\noindent \textbf{Image captioning: }
The task here is to generate a short description of an image automatically.  Image captioning models need the capability to understand an image in terms of its constituents (the objects present in the image, their attributes, the relationships between them, etc.) and to generate a proper English sentence that reflects this understanding.  
Many of the existing works are based on LSTMs or GRUs~\cite{karpathy2015deep,bottomup}.   A recent trend is to use object-level CNN features~\cite{bottomup}, and many complex frameworks have been reported in this direction.  For instance, \cite{GCNCaptioning} and  \cite{SGAECaptioning} integrated spatial object relationships via graph convolutional networks and LSTMs and scene graph architectures, respectively.  Another recent work\cite{CGANCaptioning} employed Conditional Generative Adversarial Nets to boost the captioning performance.   A recent work~\cite{aoa} proposed a complex attention mechanism that extends the object-level attention mechanism by adding another attention on top of it and achieved better results.

 Most of the best performing models for the above tasks use CNN features of the objects \& word-embeddings of the text and learn the correlation during training.  However, such an approach lacks a rich linguistic-level mapping between the modalities and may have an impaired ability to model a strong correlation.  We address this limitation by bridging the semantic gap via a simple but empirically powerful linguistically-aware approach for the attention mechanism.

\textbf{Methods using object attributes to reduce the semantic gap:}
In this section, we focus on the prior works in V-L tasks that leverage attributes from the image.   In~\cite{What_do_explicit}, the authors, trained an attribute prediction network based on a dataset-specific attribute-vocabulary that predicts a global attribute vector for the entire image that is sent to the captioning or VQA model along with the whole-image CNN feature.   One limitation of this approach is that it can only utilize the limited linguistic information available in the captions or questions while training.   To overcome this, ~\cite{wuKb} proposed to use an external knowledge-base to bring rich linguistic information to the model.   However, both of these approaches are sub-optimal in using the object attribute information and impractical in the situations demanding object-specific reasoning, e.g., Counting-VQA, since they use global image-level features.  Additionally, such approaches represent textual words (question or caption) and the attribute vector in two different learnable spaces.   Hence, while conceptually, the modalities are in the same space (textual), there is a big gap at the implementation level.   In~\cite{mlan}, the authors use a similar attribute prediction network as in~\cite{What_do_explicit,wuKb}, and represent both the question words and attributes in a common learnable space for VQA.   This approach solves the problem of representing both the modalities in a common space, but it only leverages the limited linguistic knowledge from the train set question-answer pairs.   In~\cite{fda}, the authors propose an attention mechanism for VQA that extracts keywords from the question and matches them with the attributes to find the relevant objects (using a manually selected threshold).   However, this shows inferior performance in the cases that require object visual information since it ignores the visual information of the object during the attention.

Most of the above approaches require the creation of a task- and dataset-specific attribute vocabulary that is practically not feasible.   Also, they can use only limited linguistic information from the train set since they use one-hot vectors to represent the attributes and words.   In contrast, our approach is more generic and easy to use.   We use object attributes obtained from generic object detectors along with language models to reduce the semantic gap between the modalities.   The advantage of such an approach is that it does not require any dataset-specific word vocabulary as in~\cite{What_do_explicit,wuKb,mlan} and can be used as a generic task- and model-agnostic feature.   Further, since the attributes and words are represented by pre-trained word embedding vectors, they are inherently linguistically-aware, and the model can utilize such rich linguistic information without the need for an additional knowledge base.

\section{Linguistically-aware image and text feature extraction}\label{feat_extraction}
\begin{figure}
    \centering
      \scalebox{0.9}{
    \includegraphics[width=0.8\columnwidth]{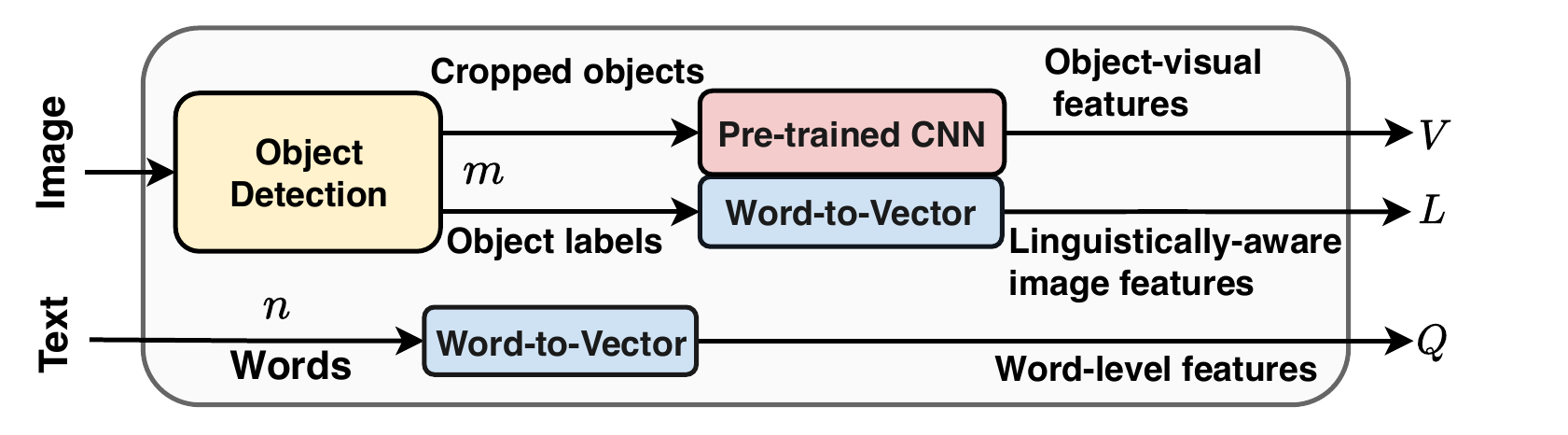}
  \caption{Linguistically-aware image and text feature extraction}
    \label{fig:feat_extraction}}
\end{figure}

In this section, we describe the proposed method to extract linguistically-aware features from a given image and text (see Fig~\ref{fig:feat_extraction}).   We build upon \cite{bottomup} that visualizes the image as a collection of object proposals.   To realize this representation format, we use a pre-trained object detector such as Faster-RCNN~\cite{fasterrcnn}, YOLO~\cite{yolo}, etc.
The visual feature extraction of these detected objects is further carried out via a pre-trained CNN (ResNet-101 \cite{resnet}) and this process generates the set of ``object-visual features'' $\textbf{\textit{V}}=\{v_1,v_2,..,v_m\}; v_i \in \mathbb{R}^{dv}$, where $m$ is the number of object regions.  

In order to establish the relationship of the object-visual features (CNN features with no linguistic information) with a semantically meaningful space, we propose to additionally represent the objects in textual space using their class labels (attributes) and extract the ``linguistically-aware image features" with pre-trained word-to-vector networks.   The pre-trained word-to-vector networks such as Glove~\cite{glove} and Bert~\cite{bert} are inexpensive and rich in making linguistic correlations (since they are already trained on a large textual corpus such as Common Crawl and  Wikipedia2014).   The ``linguistically-aware image features" provide rich object-level linguistic information to the model without training on any additional expensive textual corpus.   Formally, we represent the set of  ``linguistically-aware image features'' as $\textbf{\textit{L}}=\{l_1,l_2,..,l_m\}; l_i \in \mathbb{R}^{d_w}$, where $m$ is the number of object regions.   Similarly, we employ the same pre-trained word-to-vector representation to extract the set of word-level features from the text (i.e.,\ caption or question) as well: $\textbf{\textit{Q}}=\{q_1,q_2,..,q_n\}; q_j \in \mathbb{R}^{d_w}$, where $n$ is the length of the sequence.   The advantage of using the same word-to-vector representation for extracting both $L$ and $Q$ is that both of them get represented in the same learned space.   Such a space is built using a large text corpus and is trained such that semantically similar words have similar representations.   Thus, semantically similar ``objects'' and ``words'' will also lie close to each other in their representation space.   Hence, the mapping between the ``image'' and ``text'' will be easier and linguistically robust, and a better correlation (attention) is achieved.  In this paper, we name the attention that utilizes the linguistically-aware features $L$ and $Q$ as \textit{Linguistically-aware Attention (LAT)}.  In the following section, we show its application in various V-L tasks and models.

\section{ Application in Vision-language tasks}

This section shows the applicability of the linguistically-aware attention (LAT) in several V-L tasks: Counting-VQA, VQA, and Captioning.
In Counting-VQA, we propose a novel counting-specific VQA model based on a co-attention mechanism and tensor regression.   For VQA, we choose three best performing models, such as UpDn~\cite{bottomup}, MUREL~\cite{murel}, and BAN~\cite{ban}, into which we incorporate LAT.   In captioning, we choose the best performing object-level attention model (UpDn) from~\cite{bottomup} as the baseline and incorporate LAT.

\subsection{Counting-VQA model}

The general VQA models underperform on counting questions, mainly due to: (1) their inefficiency to handle multiple objects at a time, as they are originally tuned for a single object only, and (2) treating the task as a classification problem (classify the image and question to an answer in a predefined set of answers), that impairs the ordinal structure inherently existing in counting questions.   In light of these observations, we address Counting-VQA as a separate problem with a different objective function rather than the generic VQA problem.
We treat it as a regression problem and propose a \textit{novel Counting-VQA model} that is able to predict an intuitive count as the answer.   Such an approach is not yet reported in the literature, and our work marks the first of its kind.

\begin{figure}[t]
    \centering
      \scalebox{1}{
    \includegraphics[width=0.8\columnwidth]{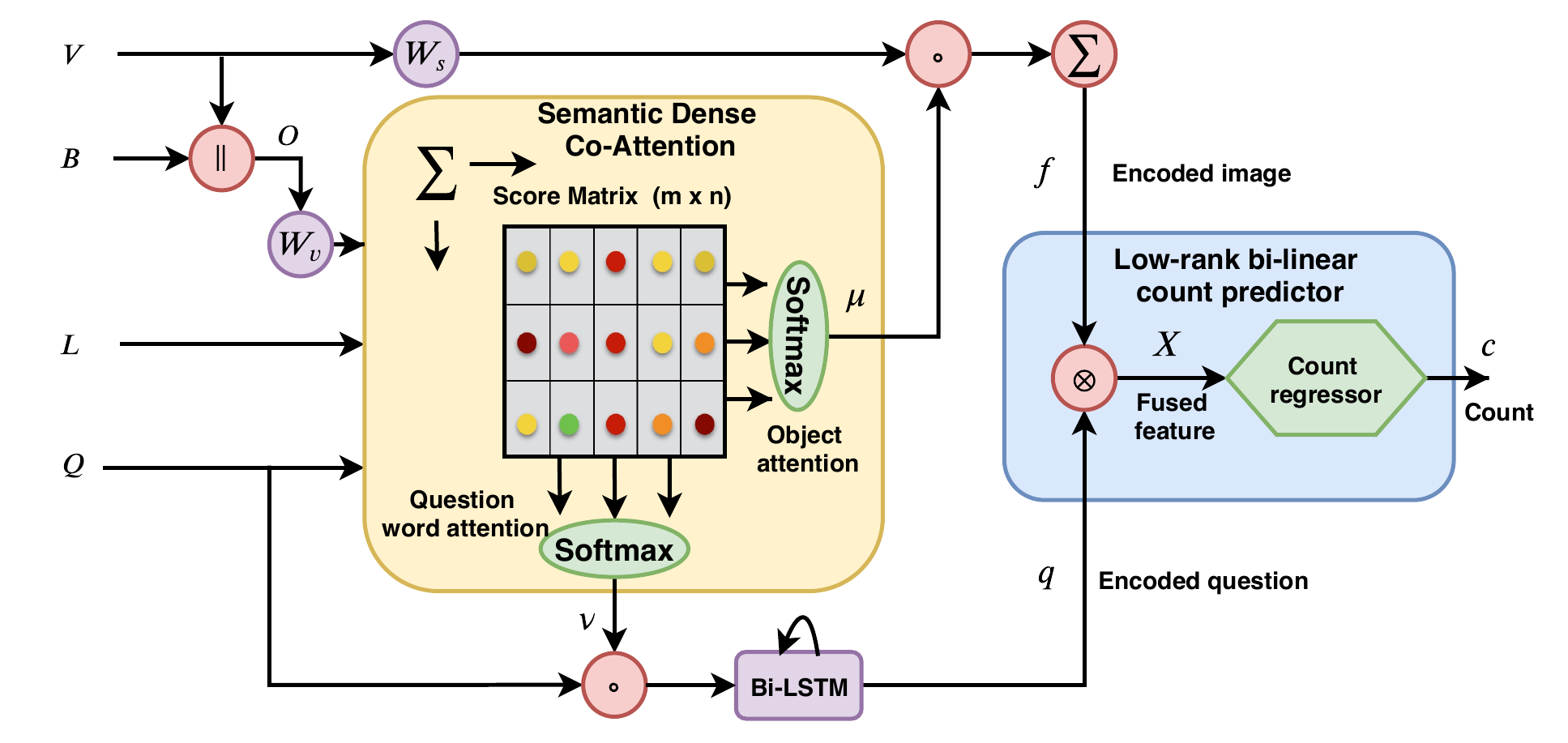}
  \caption{Counting-VQA model: The operations, $\Vert$, $\otimes$ and $\odot$ represent  concatenation, tensor product and row-wise multiplication respectively.}
    \label{fig:counting_model}}
\end{figure}

The proposed model is shown in~Fig~\ref{fig:counting_model}.   We follow an encoder-decoder approach, where the encoder encodes the image and question, and the decoder fuses them to predict the count.   The input to the model is the set of features $V$, $L$, and $Q$ (described in Sec.~\ref{feat_extraction}) and $B$, i.e., object bounding box features.  
Each entry in  $B$ corresponds to a $5D$-vector consisting of the normalized bounding box center coordinates, height, width, and area, respectively.
There are two main components in the proposed model:  (i) a semantic-level dense co-attention in the encoder and (ii) a low-rank bi-linear count predictor in the decoder.  \\
\textbf{(i) Semantic dense co-attention: }
In counting questions, certain words (e.g., nouns and adjectives) are more important than others since such questions aim to enumerate certain objects.   Hence, to make an object-specific question encoding, in addition to the object attention, we also need to attend to the question to focus on the object-relevant words.  
Inspired by this observation, we propose a co-attention mechanism (attention on both objects and question words) that we term as \textit{``Semantic dense co-attention"} since it operates at the semantic-level and finds the importance of each object and word in a co-operative manner.  The co-attention module takes the concatenated object visual and box features $O=V \Vert B $ along with $L$ and $Q$ and creates a score matrix $S$ of size $m \times n$.
Formally, it is defined as:  
\begin{equation}
    S[i,j]=W_a^T(tanh(W_v(o_i)\circ l_i \circ q_j))+b_a
    \label{score}
\end{equation}
where $S[i,j]$ represents the semantic-level correlation score between the $i^{th} $ object and $j^{th}$ question word, $\circ$ represents element-wise multiplication, $o_i,l_i$ are the $i^{th}$ object visual and box feature ($o_i \in O$) and object textual feature respectively, $q_{j}$ is the $j^{th}$ question word embedding,
 $W_v$ corresponds to the weights of a two layer network that projects $o_i \in \mathbb{R}^{d_v+5}$ to the $d_w$-D space and $W_a \in \mathbb{R}^{d_w} \text{ and }\  b_a\in \mathbb{R}$ are other learnable parameters.
Then, normalize $S$ using the $softmax$ operation along the columns and rows to get the object attention and question word attention weight vectors, $\mu \in \mathbb{R}^m$ and $\nu \in R^n$ respectively:
\begin{align}
    \mu&=softmax\Big(\sum_{j=0}^{n-1}S[:,j]\Big) \nonumber \\
    \nu&=softmax\Big(\sum_{i=0}^{m-1}S[i,:]\Big)
    \label{norm}
\end{align}
The values in $\mu$ and $\nu$ represent the relevance of the corresponding object and question word in the context, respectively.  Given  $\mu$ and $\nu $, the encoder further encodes the image and question to fixed-sized feature vectors $f$ and $q$, respectively.  The encoded image $f \in \mathbb{R}^d$ is obtained as:
\begin{align} 
    f = \sum_{i=0}^{m-1} \mu_i * (W_s^T v_i)
    \label{spatial}
\end{align}
where $W_s \in \mathbb{R}^{d_v \times d}$ is a learnable matrix to project the object visual features ($v_i\in \mathbb{R}^{d_v}$) to $d$-D space and $*$ represents multiplication.  
The encoded question $q \in \mathbb{R}^d$ is obtained by passing the weighted question word embeddings ($\nu_j * q_j$) to a Bi-LSTM and then taking the concatenated forward and backward LSTM's outputs.  
The features $f$ and $q$ are the inputs to the counting-specific decoder that is described next.  

\textbf{(ii) Low-rank bi-linear count predictor: }
The decoder learns \textit{``how to count''} from the features $f$ and $q$ by considering counting as a low-rank tensor regression problem.  The key advantage of low-rank tensor regression over traditional linear regression is that it can model the higher-order interactions between the input modalities with a lesser number of parameters~\cite{trl}.   

The count predictor first fuses $f$ and $q$ by the \textit{Tensor product (outer product)} that results in the fused feature tensor $X \in \mathbb{R}^{d \times d}$.  Then, $X$ is passed to the \textit{Count regressor} module where the count \textbf{c} is to be predicted.  Mathematically, the above steps are defined as follows (Eq.\ \eqref{trl} to Eq.\ \eqref{lowrank}): 
     \begin{align}
       	c&=round(\langle X,W_r \rangle +b_r) \nonumber\\
		\langle X,W_r \rangle &= \sum_{i=0}^{d-1}  \sum_{j=0}^{d-1}  X[i,j]*W_r[i,j] \nonumber \\
    X&=f \otimes q 
    	\label{trl}
     \end{align}
where $\otimes$ is the tensor product, $\langle \cdot,\cdot \rangle$ is the generalized inner product~\cite{trl}, and $*$ is the multiplication operation.  
The regression weight tensor $W_r \in \mathbb{R}^{d\times d}$ and the bias term $b_r \in \mathbb{R}$ needs to be learned.  Since the dimensionality of $W_r$ increases with an increase in the feature dimension $d$, which leads to a huge number of parameters to learn, we decompose $W_r$ using the low-rank Tucker decomposition.  Given rank $k$, the Tucker decomposition decomposes $W_r$ into a core tensor $T_c \in \mathbb{R}^{k \times k}$ and factor matrices $\{W_q,W_f\} \in \mathbb{R}^{d\times k}$ as: 
\begin{align}
    W_r \approx W_q T_c W_f^T
    \label{tucker}
\end{align}
Applying Eq.\ \eqref{tucker} to Eq.\ \eqref{trl} gives the final representation of our count predictor as in Eq. \eqref{lowrank}, where $T_c\in \mathbb{R}^{k \times k}$, $\{W_q,W_f\} \in \mathbb{R}^{d\times k}$ and $b_r \in \mathbb{R}$ are learnable parameters.  Since the output of the regression step is a real number, we take the rounded integer value to get the count (rounding is applied only during testing):
\begin{align}
    c=round(\langle X,(W_q T_c W_f^T) \rangle +b_r)
    \label{lowrank}
\end{align}

\subsection{Incorporating LAT into various VQA models}\label{vqa_model}

In VQA, we study the effect of LAT on three recent best performing models: UpDn~\cite{bottomup}, MUREL~\cite{murel}, and BAN~\cite{ban}.  All of these models are based on the object-level CNN representations proposed in~\cite{bottomup}.  In the following, we explain the changes applied to the baseline models in order to use LAT.

\subsubsection{Incorporating LAT into the UpDn baseline: }

\begin{figure}
    \centering
      \scalebox{1}{
    \includegraphics[width=0.95\columnwidth]{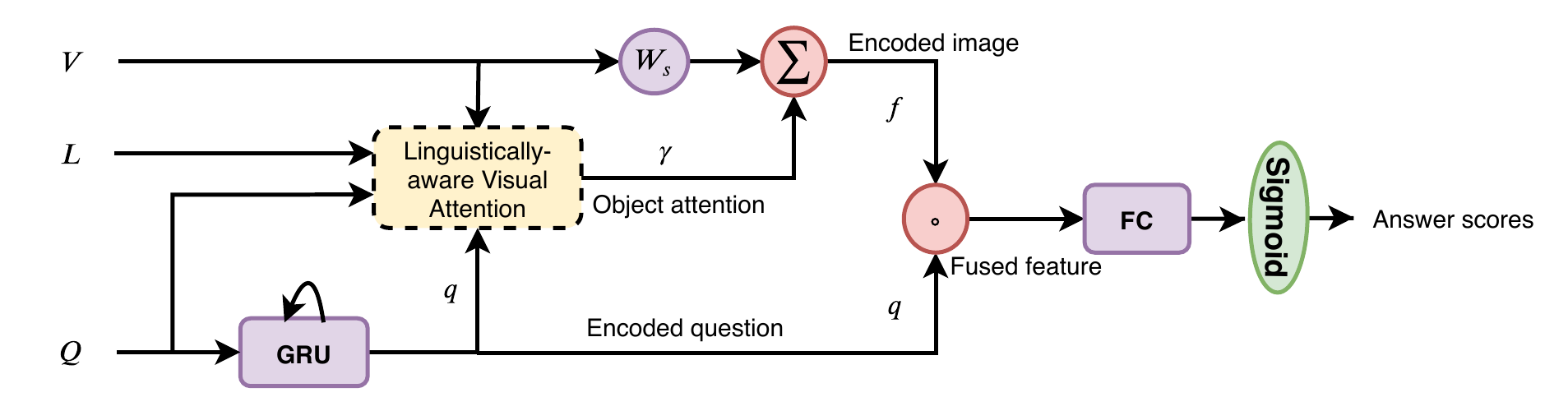}
    \caption{UpDn VQA model with LAT:  $\circ$ is element-wise multiplication.  The dotted box represents our addition.}
    \label{fig:vqa_model}}
\end{figure}

This model (UpDn) is purely based on the object-level visual attention proposed in~\cite{bottomup}.  We use LAT in the visual attention part of the baseline model.  The model with LAT is shown in Fig~\ref{fig:vqa_model}.  
As in the baseline~\cite{bottomup}, the question is encoded by sending the word-level features ($Q$) to a GRU resulting in the question context vector $q \in \mathbb{R}^d$.  
 The image is encoded using the linguistically-aware visual attention mechanism.  
 We use the \textit{gated tanh} non-linear function $f_{ga}: x \in \mathbb{R}^p \rightarrow y \in \mathbb{R}^r$ as in \cite{bottomup} in our visual attention, as it shows better performance for this problem than traditional $ReLU$.  It is defined as:
\begin{align}
y&=tanh(Wx+b) \circ sigmoid(W^\prime x+b^\prime)
\label{gtanh}
\end{align}
where $\{W,W^\prime\} \in \mathbb{R}^{r \times p}$ and $\{b, b^\prime\} \in \mathbb{R}^r$ are learnable parameters.

\textbf{Linguistically-aware visual attention:}
We follow a similar attention formulation as in the baseline~\cite{bottomup} to make a fair comparison.  In addition to the object-visual features $V$ and the question context vector $q$, 
 we incorporate the features $L$ and $Q$ to provide linguistic-awareness.  The visual attention module outputs a vector $\gamma \in \mathbb{R}^m$, that represents the relevance of each of the $m$ objects in the context.   
Formally, our visual attention mechanism is described as: 
\begin{align}
\gamma &= softmax(s) \nonumber \\
s_i&= W_v^Tf_v([v_i,q])+\sum_{j=1}^n(W_l^Tf_l(l_i \circ q_j)+b_l)
    \label{vqa:att}
\end{align}
where $s \in \mathbb{R}^m$ is the score vector with $s_i \in \mathbb{R}$ is the score for the $i^{th}$ object, $\gamma \in \mathbb{R}^m$ is the attention weight vector, $f_l :\mathbb{R}^{dw} \rightarrow \mathbb{R}^{dw}$, $f_v: \mathbb{R}^{dv} \rightarrow \mathbb{R}^{dw}$ are \textit{gated tanh} functions as described in Eq.  \eqref{gtanh},  and $\{W_l, W_v\} \in \mathbb{R}^{d_w}$, $b_l \in \mathbb{R}^{d_w}$, are learnable parameters.
The encoder uses the attention weight vector $\gamma$ and encodes the image by taking the weighted sum of the object visual features $V$ as described in Eq. \eqref{spatial} (using $\gamma_i$ instead of $\mu_i$), resulting in the encoded image $f \in \mathbb{R}^d$.  

As in the baseline~\cite{bottomup}, the decoder $D(f,q)=sigmoid(W_o^T(f \circ q))$ fuses $f$ and $q$ by element-wise multiplication and predicts the answer scores by a fully connected layer (FC) with the \textit{sigmoid} function, 
where $W_o \in \mathbb{R}^{d \times d_o}$
is a learnable parameter, and $d_o$ is answer vocabulary size.

\subsubsection{Incorporating LAT into the MUREL baseline:}

 MUREL~\cite{murel} is a non-attention based model.  The question word features ($Q$) are given to a GRU, and the final hidden state vector is taken as the encoded question $q$.  Then, the object-visual features $V$ along with $q$ are passed to the ``MUREL cell".  Inside this cell, each of the object-visual features ($v_i\in V$) are bi-linearly fused with $q$ resulting in the feature vector $s_i$ corresponding to $v_i$.  A global max-pooling operation is performed over the $s_i$ features, and the resulting vector is considered the encoded image $s$.  Then the encoded image $s$ and question $q$ are bi-linearly fused to predict the answer.

\begin{figure}[t]
    \centering
    \footnotesize
    \includegraphics[width=0.9\columnwidth]{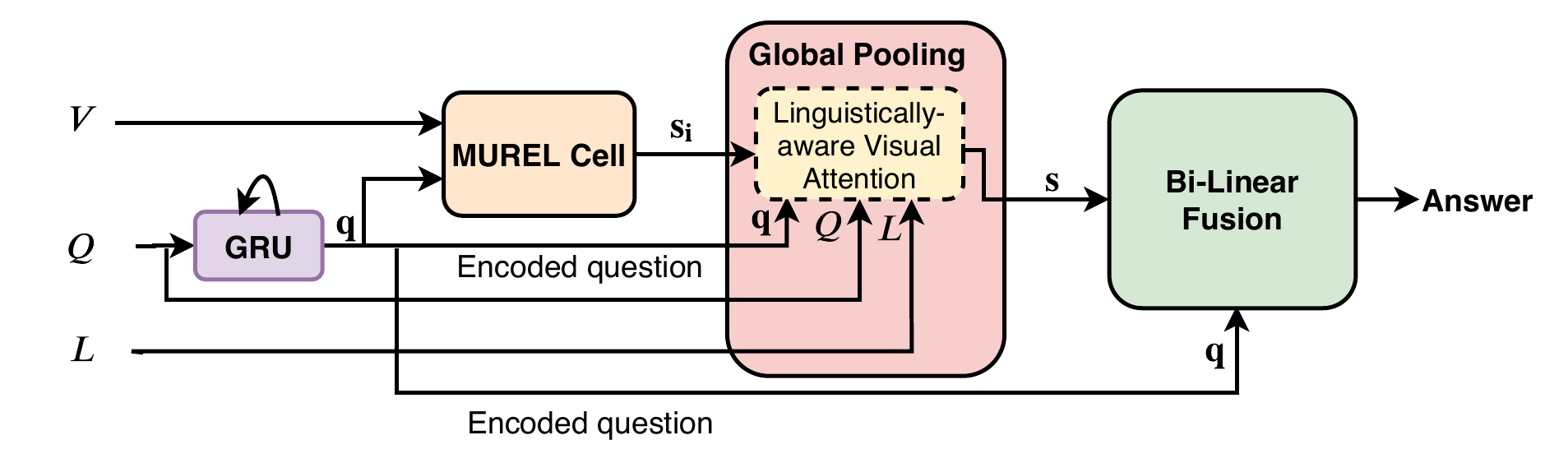}
  \caption{MUREL VQA model with LAT.  The dotted box represents our addition.}
  \label{fig:murel}
\end{figure}

We adapt LAT into the MUREL model, as shown in Fig.~\ref{fig:murel}.
Instead of the global max-pooling operation in the original model, we have applied weighted average pooling.  The weights are defined by the ``Linguistically-aware visual attention" weight vector $\gamma$, formulated in Eq.~\eqref{vqa:att}.  

\subsubsection{Incorporating LAT into the BAN baseline:}

The BAN~\cite{ban} is based on bi-linear co-attention between the object-visual and question word features ($V$ and $Q$).  The question word features are passed through a GRU, and all the hidden state vectors are taken.  These are denoted by $\hat Q \in R^{n \times d}$, where $n$ is the number of words, and $d$ is the size of the GRU hidden state.  Then, both $\hat Q$ and $V$ are given to the bi-linear attention network (BAN) to get the combined vector $f_v \in R^C$.  Formally, it is denoted as $f_v=BAN(\hat Q,V;\mathcal{A}_1)$, 
where $\mathcal{A}_1$ is the bi-linear attention map between $\hat Q$ and $V$. The $f_v$ is further passed to a Multi-layer Perceptron (MLP) to predict the answer.

\begin{figure}[t]
    \centering
    \footnotesize
    \includegraphics[width=0.7\columnwidth]{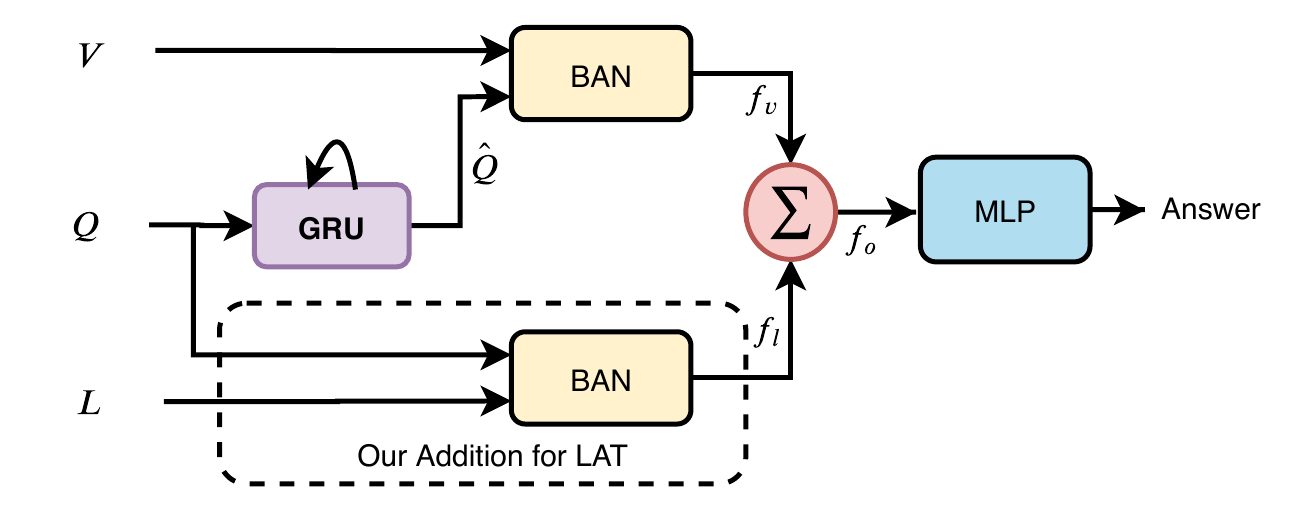}
  \caption{BAN VQA model with LAT.  The dotted box represents our addition.}
  \label{fig:ban}
\end{figure}

We have incorporated LAT into the BAN model by adding co-attention maps between the linguistically-aware image features $L$, and the question word features $Q$.  An overview of the BAN model with LAT is shown in Fig.~\ref{fig:ban}.  
In another perspective, this addition can be considered as a bi-linear linguistically-aware co-attention since it is between the linguistically-aware features $L$ and $Q$.   Formally, the above operations are described as:
\begin{align}
f_o= fv+f_l=BAN(\hat Q,V;\mathcal{A}_1)+BAN(Q,L;\mathcal{A}_2)
    \label{eq:ban}
\end{align}
where $f_v$ is the bi-linear joint representation as in the original model~\cite{ban}, $f_l$ is from the LAT adaptation, and $f_o$ is the final joint representation which is then passed to the MLP to predict the answer probabilities.

\subsection{Incorporating LAT into the UpDn captioning model}
\begin{figure}[t]
    \centering
    \includegraphics[width=0.75\columnwidth,height=8.5cm]{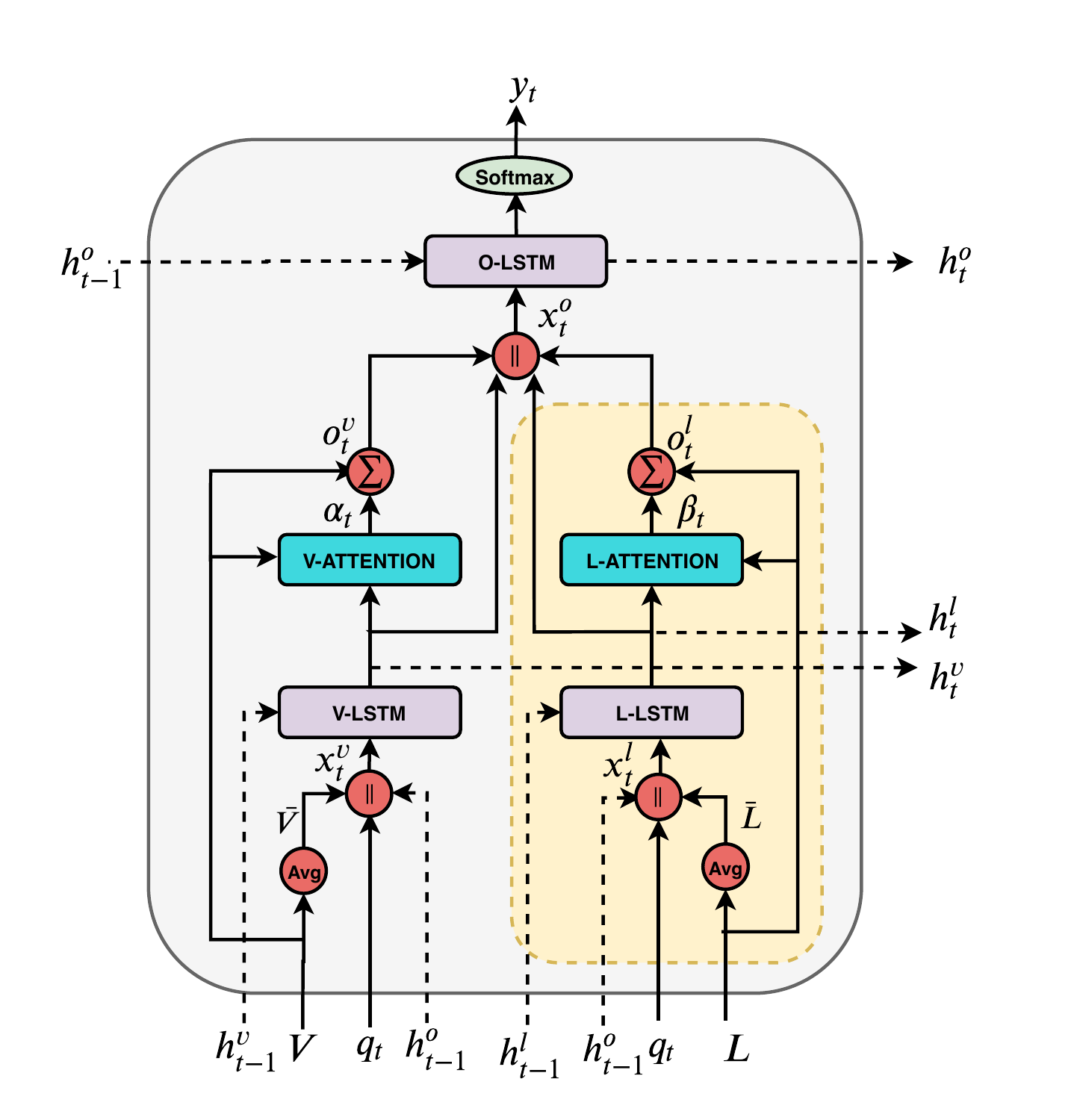}
  \caption{UpDn captioning model LSTM cell with LAT, at time $t$: Dotted lines indicate hidden states.  The operations $\Vert$, $\sum$ and $Avg$ are the Concatenation, Weighted sum and Average respectively.  Dotted box represents our addition.}
  \label{fig:cap}
\end{figure}

Similar to the baseline, we follow the LSTM based encoder-decoder approach.  
We use the notation $h_t=\text{LSTM}(x_t,h_{t-1})$ for an LSTM cell, where $x_t$ is the input at time $t$, and $h_{t-1}$ and $h_t$ are the previous and current cell outputs, respectively.
We incorporate LAT into the baseline model~\cite{bottomup}, as described below.  For ease of explanation, we consider the details of the LSTM cell at time $t$ (see Fig.~\ref{fig:cap}).  
It consists of three layers: (1) An input layer, (2) an attention layer, and (3) an output layer.  The first two layers together constitute the encoder, where the current caption word, the previously generated caption context vector, and object visual and linguistic information are aggregated to fixed-sized context vectors.   The output layer is the decoder, that predicts the next caption word probabilities from the encoded vectors.  

\textbf{Input layer:}
This layer consists of two LSTM cells viz., V-LSTM and L-LSTM.
The V-LSTM is analogous to the Top-Down Attention LSTM in the baseline model~\cite{bottomup}.  
It takes the average object-visual features $\bar{V}=\frac{1}{m}\sum_iv_i$, current caption word feature $q_t$ and the context vector for the caption generated so far $h^o_{t-1}$ and outputs the object-visual context vector  $h^v_t\in \mathbb{R}^{d_e}$.  In addition to this, we incorporate another LSTM cell called L-LSTM that takes the average linguistically-aware image features  $\bar{L}=\frac{1}{m}\sum_il_i$, current caption word feature $q_t$ and the context vector for the caption generated so far $h^o_{t-1}$ and outputs the object-textual context vector  $h^l_t\in \mathbb{R}^{d_e}$.
Mathematically, the above operations are described as:
\begin{align}
    h^v_t=\text{V-LSTM}(x^v_t,h^v_{t-1})\;\;; \; \; 
    x^v_t=[h^o_{t-1},\bar{V},q_t] \nonumber \\  
    h^l_t=\text{L-LSTM}(x^l_t,h^l_{t-1}) \;\;; \; \;
    x^l_t=[h^o_{t-1},\bar{L},q_t]
\end{align} 

\textbf{Attention layer:}
This layer encodes the object visual and textual features ($V$ and $L$) into fixed-sized vectors using attention.  
It consists of two attention modules, the
object-visual attention ``V-Attention" (analogous to the ``Attend" module in the baseline~\cite{bottomup}) and the linguistically-aware attention ``L-Attention" (our contribution for LAT).
The V-Attention and L-Attention modules output the attention weights $\alpha_t$ for $V$ and $\beta_t$ for $L$ respectively, at time $t$.  Formally, the two modules are defined as:
\begin{align}
      \alpha_t=softmax( (tanh(  \mathbf{V}W^v + (\mathbf{1} h_t^{v^T})W_h^v)W^v_a))
       \nonumber\\
       \beta_t=softmax((tanh(\mathbf{L}W^l + (\mathbf{1} h_t^{l^T})W_h^l)W^l_a))
\end{align}
where $\mathbf{1}$ is a column vector of length $m$ consisting of all ones, $\mathbf{V} \in \mathbb{R}^{m \times d_v}$, $\mathbf{L} \in \mathbb{R}^{m \times d_w}$ are the matrix representations of the input features $V$ and $L$ respectively, $\{W^v_a, W^l_a \} \in \mathbb{R}^d$, $W^v \in \mathbb{R}^{d_v \times d}$, $W^l \in \mathbb{R}^{d_w \times d}$ and $\{W_h^v, W_h^l\} \in \mathbb{R}^{d_e \times d}$ 
are learnable parameters.  Then the fixed sized object-visual encoding $o^v_t \in \mathbb{R}^{d_v}$ and object-textual encoding $o^l_t \in \mathbb{R}^{d_w}$ are obtained as in Eq.   \eqref{eq:cap_encode}, where $\alpha_t[i]$ and $\beta_t[i]$ are the attention weights for the $i^{th}$ object-visual feature $v_i$ and object-textual feature $l_i$ at time $t$, respectively.
\begin{align}
o^v_t&=\sum_{i=1}^{m} \alpha_t[i] v_i &
o^l_t&=\sum_{i=1}^{m} \beta_t[i] l_i 
\label{eq:cap_encode}
\end{align}

\textbf{Output layer:}
This is the decoder that predicts the probability scores for the next caption word having seen the partial caption generated so far.  It consists of the O-LSTM (analogus to the ``Language LSTM" in the baseline~\cite{bottomup}), that takes $x^o_t=[o^v_t,o^l_t,h^v_t,h^l_t]$ and the previous cell hidden state vector $h^o_{t-1}$ as inputs and then outputs a vector $h^o_t \in \mathbb{R}^{d_o}$.  It is further passed to a $softmax$ function to get the next word probabilities $y_t \in \mathbb{R}^{d_o}$ ($d_o$: size of output vocabulary).   

\section{Experimental Setup}
\noindent \textbf{Feature representations}:
In all the tasks and models, we use the features provided by~\cite{bottomup}, i.e., the fixed size ($m=36$) object-level ResNet-101~\cite{resnet} features extracted using Faster-RCNN~\cite{fasterrcnn} trained on Visual Genome~\cite{genome}, as the set of object-visual features $V$ ($d_v=2048$).  
Then, we extract the object attributes corresponding to each of the $m$ objects.  In the Counting-VQA model, we also use YOLOv3 pre-trained on MSCOCO as the object detector and extract the CNN features using ResNet-101.  As the word-to-vector representation, we use pre-trained  Glove~\cite{glove} (for Counting-VQA and VQA; $d_w=300$) and Bert~\cite{bert} (for Image captioning; $d_w=1024$).  For various experiments, we have used other features as well, which will be explained in the respective sections.  

\noindent \textbf{Datasets and Implementation details}:\\ 
\textbf{(i) Counting-VQA: } 
We use the following counting-specific VQA datasets:
\textbf{\textit{HowManyQA: }} This dataset~\cite{irlc} is the counting-specific union of two large VQA datasets: Visual Genome~\cite{genome} and VQAv2~\cite{vqa2}.  
It includes $83642$ (train), $17714$ (validation) and $5000$ (test) question-answer pairs.  
\textbf{\textit{TallyQA: }}
This dataset~\cite{tallyqa} consists of a large test set categorized into two: ``simple" ($22991$ question-answer pairs) and ``complex" ($15598$ question-answer pairs), where the former contains basic object enumeration questions, and the latter requires more reasoning.  
\textbf{\textit{Count-QA (VQAv1)}} and \textbf{\textit{Count-QA (COCO-QA): }} These datasets~\cite{everyday} contain counting questions from VQAv1~\cite{vqa1} and COCO-QA~\cite{cocoqa} with $1774$ and $513$ question-answer pairs respectively.

In the count regression step, we use the low-rank value $k=11$.  After all the linear layers, we use \textit{Batch Normalization} and \textit{ReLU}.  We train the model on the HowManyQA dataset using \textit{Smooth L1 loss}~\cite{fastrcnn} and \textit{Adam} optimizer with $learning\ rate=0.0005$ and $batch\ size=8$.

\noindent \textbf{(ii) VQA:} We use the VQAv2 dataset \cite{vqa2} containing $1.1$M questions with $11.1$M answers for training and evaluating all the models.
As the evaluation metric, we use the standard VQA accuracy~\cite{vqa1}.  
For ease of implementation, we limit the question length to 14 and use all the answers occurring more than eight times as the answer vocabulary as in~\cite{bottomup,ban}.  In all the models, we use a similar training and optimization settings as in the respective models~\cite{bottomup,murel,ban}.

\noindent \textbf{(iii) Image captioning:}
We use the ``Karpathy splits" of  MSCOCO captions dataset~\cite{MSCOCO_Captions} that contains $113K$ training images and $5K$ images in the validation and test set each (five captions per image).  
We use the caption output vocabulary of size $d_o=9487$ as in \cite{bottomup}.  
We evaluate our model using the standard image captioning evaluation metrics such as BLEU, METEOR, ROUGE-L, CIDEr, and SPICE.  We train our model using Cross-Entropy loss and CIDEr optimization using the $Adam$ optimizer with \textit{batch\_size=$16$} and \textit{learning\_rate=$0.0005$}.  

\begin{table}[t]
    \centering
    \footnotesize
    \begin{tabular}{cc}
    \begin{tabular}{p{0.2\linewidth} c}
    \toprule
         Model & HowManyQA \\
         \midrule
         detect~\cite{everyday}&3.66 \\
         MUTAN~\cite{mutan}&2.93 \\ 
         UpDn~\cite{bottomup}&2.64\\
         Counter~\cite{iclr2}&2.59\\
         IRLC~\cite{irlc}&2.47\\
         RCN~\cite{tallyqa}&2.35\\
         \midrule
         \multicolumn{2}{c}{\textbf{Ours}}\\
         \midrule
         F-genome&1.62\\
         YOLO&\textbf{1.56}\\
         YOLO+BERT&\textbf{1.52}\\
         \bottomrule
    \end{tabular}
    &\begin{tabular}{p{0.2\linewidth}p{0.08\linewidth}p{0.08\linewidth}}
         \toprule
         &\multicolumn{2}{c}{TallyQA}
         \\
         Model & Simple & Complex \\
         \midrule
         detect~\cite{everyday}&2.08&4.52\\
         MUTAN~\cite{mutan}&1.51&1.59 \\
         Counter~\cite{iclr2}&1.15&1.58\\
         RCN~\cite{tallyqa}&1.13&1.43\\
         \midrule
         \multicolumn{3}{c}{\textbf{Ours}}\\
         \hline
         F-genome&N/A&1.02 \\
         WO VG&\textbf{0.97}&1.03 \\
         YOLO&N/A&\textbf{0.94} \\
         YOLO+BERT&N/A&\textbf{0.93}\\
         \bottomrule
    \end{tabular}
             \\ \\
         \multicolumn{2}{c}{
         \begin{tabular}{lcc}
    \toprule
        Model & VQAv1& COCO-QA \\
          \midrule
          MCB~\cite{mcb} &  3.25&-\\
          D-LSTM~\cite{deeperlstm}&2.71&-\\
          detect~\cite{everyday}&2.72&2.59\\
          ens~\cite{everyday}&1.80&1.4 \\
          \midrule
          \makecell[l]{\textbf{Ours} (F-genome)}&1.01&0.94\\
          \makecell[l]{\textbf{Ours} (YOLO)}&\textbf{0.98}&\textbf{0.92} \\
          \textbf{Ours}(YOLO+BERT)&\textbf{0.96}&\textbf{0.9}\\
          \bottomrule
    \end{tabular}}
    \\
    \end{tabular}
    \caption{Counting-VQA results on HowManyQA, Count-QA splits of VQAv1 and COCO-QA and TallyQA simple and complex test sets.  All are RMSE values.  \textbf{Lower} is better.}
   \label{tab:countqa}
\end{table}

\section{Results and Discussions}

In this section, we show the results of various experiments that we carry out using LAT on the Counting-VQA model (Sec.  \ref{Sec:countVQA}).  In VQA and Captioning (Sec.  \ref{VQAandCAP}), we show the effect of LAT on various recent best performing models.
\subsection{Counting-VQA}
\label{Sec:countVQA}
\textbf{Comparison with state-of-the-art methods: }
The quantitative comparison of our model
with existing approaches on the HowManyQA, TallyQA (Simple and Complex), and Count-QA (VQAv1 and COCO-QA) datasets are shown in Table~\ref{tab:countqa}.  We show the cases of two types of image features (YOLOv3~\cite{yolo} and F-genome~\cite{bottomup}) and language models (Glove~\cite{glove} and BERT~\cite{bert}).  The F-genome features are the same features as used in prior works, i.e., features from~\cite{bottomup}.
The TallyQA-simple and the HowManyQA train set may contain common questions from Visual Genome~\cite{genome}, hence to make a fair comparison, we also show the results without using such questions (WO VG in Table~\ref{tab:countqa}).  
We can see that our model outperforms all the existing models in all the datasets.  

Comparing the performance among the image features, the YOLO features work better than the F-genome features.  To further analyze this, we performed various experiments as detailed in Sec~\ref{count:additional_Exp}.  In the language model side, BERT outperformed Glove, as expected.

\begin{figure}[t!]
     \centering
     \footnotesize
     \setlength\tabcolsep{0.3pt}
     \begin{tabular}{cccc}
          \includegraphics[width=0.2\columnwidth,height=2cm]{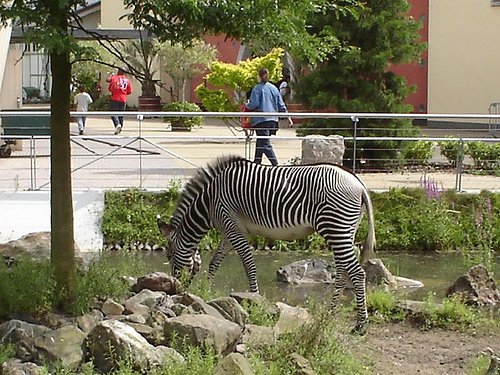}& \includegraphics[width=0.2\columnwidth,height=2cm]{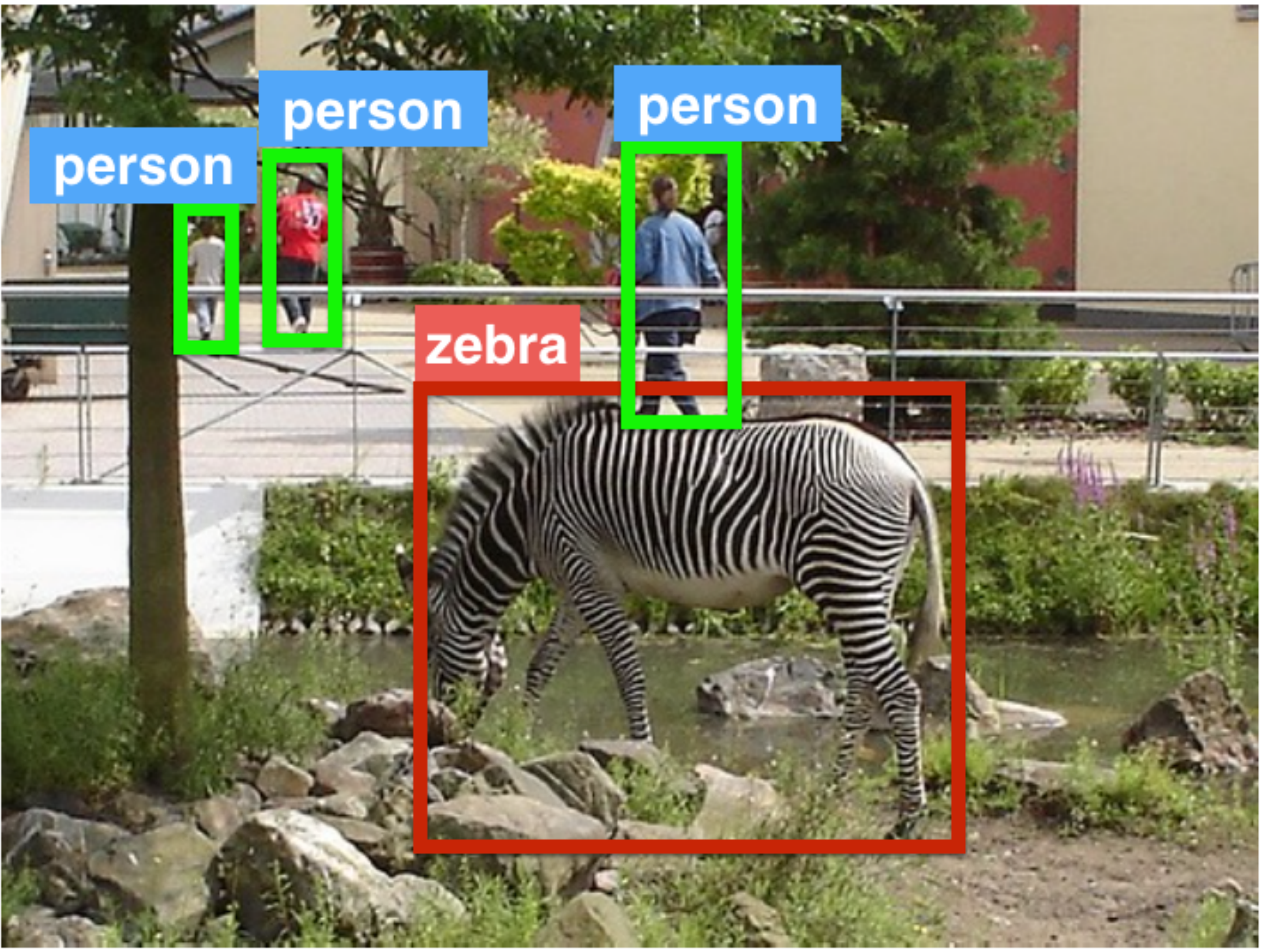}&
          \includegraphics[width=0.2\columnwidth,height=2cm]{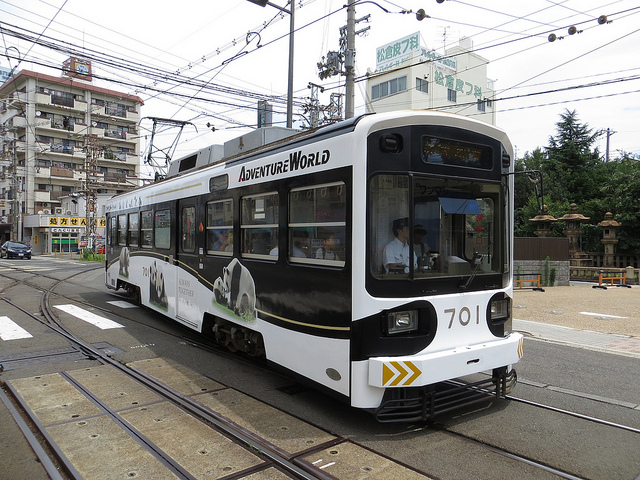} &
\includegraphics[width=0.2\columnwidth,height=2cm]{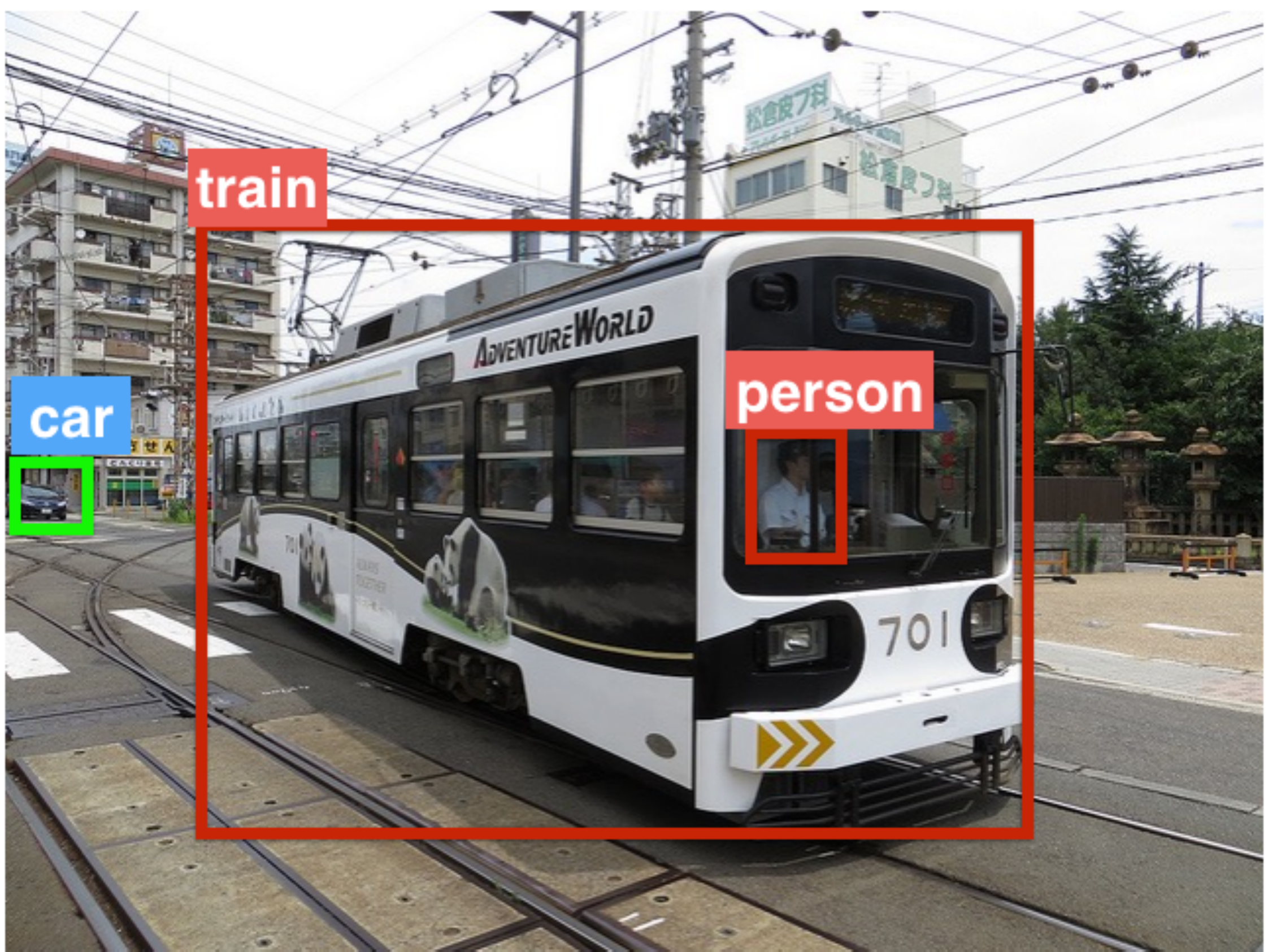}\\
   \makecell[c]{How many people \\appear in the \\ picture \;\; \textbf{GT:} 3}&
 \makecell[c]{\textcolor{blue!30!white}{How} \textcolor{blue!70!white}{many} \textcolor{blue!100!white}{people}\\ \textcolor{blue!60!white}{appear} \textcolor{blue!20!white}{in} \textcolor{blue!30!white}{the}\\ \textcolor{blue!50!white}{picture} \;\;
 \textbf{Pred:} 3} &\makecell[c]{How many sedans \\are in the picture\\ \textbf{GT:} 1} &\makecell[c]{\textcolor{blue!30!white}{How} \textcolor{blue!20!white}{many} \textcolor{blue!100!white}{sedans} \\ \textcolor{blue!60!white}{are} \textcolor{blue!50!white}{in} \textcolor{blue!20!white}{the} \textcolor{blue!70!white}{picture}
 \\\textbf{Pred:} 1} \\
 \includegraphics[width=0.2\columnwidth,height=2cm]{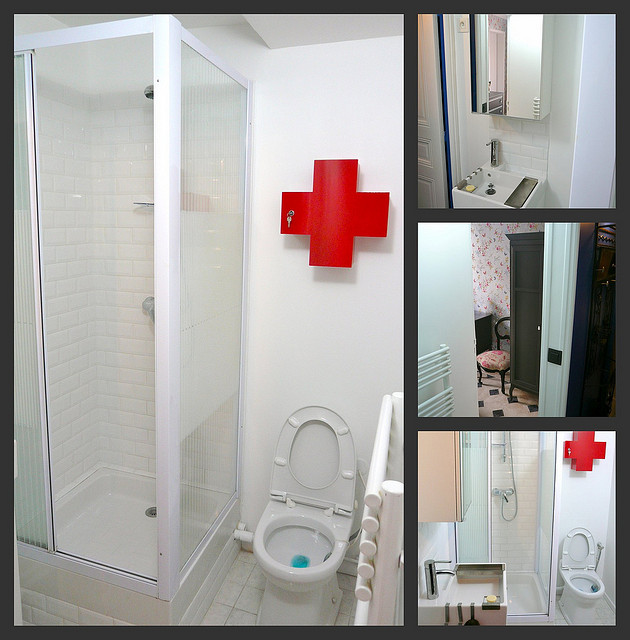}& \includegraphics[width=0.2\columnwidth,height=2cm]{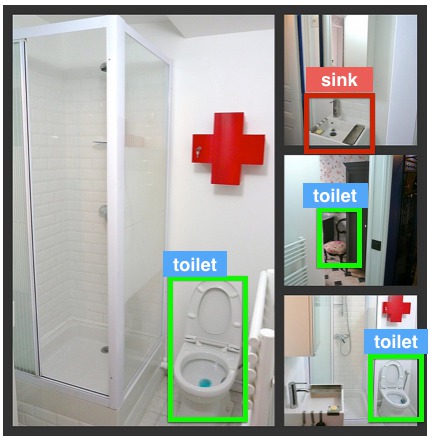}&\includegraphics[width=0.2\columnwidth,height=2cm]{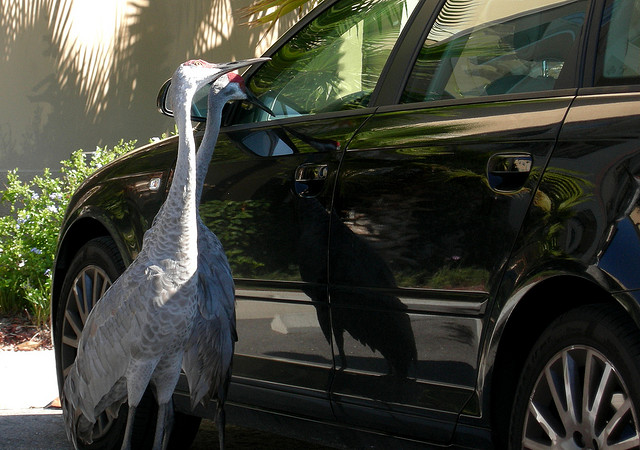} &
\includegraphics[width=0.2\columnwidth,height=2cm]{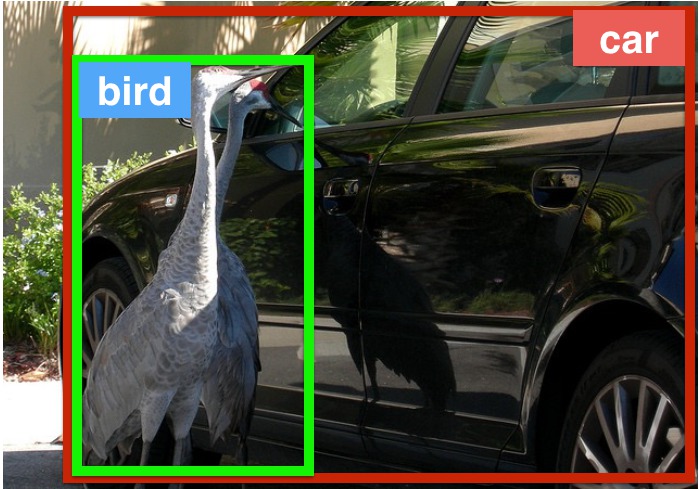}
 \\
 \makecell{How many toilets \\ \textbf{GT: }2} &\makecell{\textcolor{blue!30!white}{How} \textcolor{blue!20!white}{many} \textcolor{blue!100!white}{toilets}\\  \textbf{Pred:} 3
}& 
  \makecell{How many birds are\\ in the photo \\ \textbf{GT: }2} & \makecell{\textcolor{blue!20!white}{How} \textcolor{blue!40!white}{many} \textcolor{blue!100!white}{birds}  \textcolor{blue!10!white}{are}\\ \textcolor{blue!40!white}{in} \textcolor{blue!50!white}{the}
  \textcolor{blue!60!white}{photo} \\ \textbf{Pred: }1}\\ 

    \end{tabular}
    \caption{Counting-VQA results: Green box denotes high attention ($\mu_i \ge 0.1$) and red box denotes low attention ($\mu_i < 0.1$) on the objects.  The color intensity in question words indicates attention weights.  The bottom row shows examples of failure cases.  Best viewed in color.  }\label{fig:countqa_qual}
 \end{figure}

 \textbf{Qualitative results:} In Fig~\ref{fig:countqa_qual}, we show some qualitative results, where the first pair involves a basic linguistic understanding to correlate ``people" with the object label ``person".  The second pair is a more challenging scenario demanding high-level linguistic awareness, wherein the model has to relate ``sedan" to ``car," rather than to ``person" or ``train." 
  From the qualitative results, we can see that our model can focus on the  relevant objects and question words and predict the count accurately.  From the failure cases (bottom row in Fig~\ref{fig:countqa_qual}), we can see that our model's performance is tightly coupled with the object detector predictions.  To analyze the effect of the object-detector quality on the performance, we perform an experiment as detailed in Sec.~\ref{count:additional_Exp} (i).  
  
\subsubsection{Ablation studies }
To study our Counting-VQA model in detail, we perform the following ablation studies, and the results are shown in Table~\ref{tab:count_ab}.  All of the following experiments are carried out with the YOLOv3  and Glove features.

\begin{table}[t]
    \centering
    \footnotesize
    \begin{tabular}{lcc}
\toprule
\makecell[c]{Model}& \makecell[c]{HowManyQA} &\makecell{TallyQA-complex}\\
\midrule
\makecell[l]{No co-attention} & 1.86  &1.23\\
\makecell[l]{No L feature}&1.80 &1.20\\
\makecell[l]{No V and B features} & 1.67&1.10\\
\makecell[l]{No B feature}&1.58&0.96\\
\makecell[l]{Linear regression}  & 1.74 &1.17 \\
\midrule
\makecell[l]{\textbf{Full model}}&\textbf{1.56}&\textbf{0.94}\\
\bottomrule
\end{tabular}
    \caption{Counting-VQA: Ablation study results (RMSE).  }
    \label{tab:count_ab}
\end{table}

\textbf{\textbf{(i) No co-attention: }}In this test, we do not attend to the question words.  It is analogous to every word getting equal importance. The performance decreases (from 1.56 to 1.86 in HowManyQA and 0.94 to 1.23 in TallyQA-complex), since the model doesn't focus on the object-relevant words such as nouns and adjectives, and the irrelevant words contribute noise to the regressor.  This result validates our claim that counting benefits from semantic-level co-attention.

\textbf{(ii) Contribution of different features:}
To understand the contribution of the features $V$, $B$ and $L$ in the ``Semantic dense co-attention", we perform the following experiments:
\textbf{\textit{a) No L feature: }} Here, we remove the linguistically-aware image features ($L$) from the co-attention stage.  We observed a reduction in the performance (from 1.56 to 1.80 in HowManyQA and 0.94 to 1.20 in TallyQA-complex).  
\textbf{\textit{b) No V and B features: }} In this experiment, we keep the $L$ feature and remove the object-visual (V) and box (B) features from the co-attention.  Specifically, we remove the $W_v(o_i)$ term from Eq.\ \eqref{score}.  We observe a small reduction in the performance (from 1.56 to 1.67 in HowManyQA and 0.94 to 1.10 in TallyQA-complex).
\textbf{\textit{c) No B feature: }}We also experiment by removing only the object box features ($B$) from the co-attention (we used $W_v(v_i)$ instead of $W_v(o_i)$ in Eq.  \eqref{score}).   We observed a reduction in performance (1.56 to 1.58 in HowManyQA and 0.94 to 0.96 in TallyQA-complex).

From the above experiments (a,b, and c), we infer that the $L$ features can serve as a bridge between the modalities to make a better correlation.  However, some questions may require the appearance and relative position of the objects (e.g., How many yellow bananas are near the glass?).  In such cases, the object-visual and box features are also required for better performance.

\textbf{(iii) Linear regression:}  In this experiment, we fuse $f$ and $q$ by element-wise multiplication and use linear regression to predict the count, instead of the proposed count predictor.  The performance decreases (1.56 to 1.74 in HowManyQA and 0.94 to 1.17 in TallyQA-complex).  This result shows the ability of our low-rank bi-linear count predictor to predict the count more accurately.  

\subsubsection{Other experiments on the Counting-VQA model}\label{count:additional_Exp}
\textbf{(i) Quality of the object detector:}  To understand the influence of the object detector quality on counting performance, we experiment using the manually annotated objects vs. object detector predictions.  We use the Visual Genome~\cite{genome} question-answers (36100 pairs) from the HowManyQA train set for this experiment.  We then train two separate models with manually annotated objects (from Visual Genome) and YOLOv3 predictions and compare the performance on the HowManyQA test set (5000 question-answer pairs).  We observe that the predicted objects achieve similar performances (RMSE: 1.85) as in manual annotation (RMSE: 1.84), showing that the current object detectors can predict and classify the object regions with sufficient accuracy, thanks to the advancements in the object detection research.

\textbf{(ii) Changing the object detector:} To understand how the model performs with features from various object-detectors, we experiment with two more object-detectors apart from YOLOv3, i.e., the CenterNet~\cite{centernet} and Faster-RCNN~\cite{fasterrcnn}, both trained on MSCOCO dataset.  Table~\ref{tab:count-feat} shows the results (CenterNet and Faster-COCO).  We observe a similar performance in all cases.  We infer that changing the object-detector does not affect the performance much.  

\textbf{(iii) Effect of the number of classes in the object detector:} 
To study how the model behaves upon adding more classes to the object detector, we experiment with two variants of Faster-RCNN~\cite{fasterrcnn};
\textit{Faster-COCO:} Trained on MSCOCO (80 classes) and \textit{Faster-VG: } Trained on Visual Genome (3000 classes).  In both cases, we extract the objects in the same way as YOLOv3, i.e., with an object confidence threshold of 0.7.
In both cases, we got a similar performance as with YOLOv3 (see Table~\ref{tab:count-feat}).  
We observe that the addition of classes to the object detector did not contribute much to the performance.   This is because Faster-COCO, even though trained on a lesser number of classes, predicts the unknown class objects to some semantically similar known classes, for e.g., ``man" and ``woman" are predicted as ``person," ``jar" is predicted as ``bottle," etc.  Since they are semantically similar, the corresponding word embeddings also behave the same, and the performance is not affected much.    

\textbf{(iv) Effect of the object confidence threshold on performance: }
Another observation from the above experiment is the effect of the object confidence threshold in the object detector.  This threshold denotes the minimum confidence value needed to consider a region as an object.  
In Table~\ref{tab:countqa}, we observed a small performance gap in the counting model with F-genome features~\cite{bottomup} (extracted using Faster-RCNN trained on 1600 Visual Genome classes), but we did not see such a gap with the Faster-COCO and Faster-VG features (see Table~\ref{tab:count-feat}).  Since the F-genome features correspond to the top $36$ object predictions and may include objects with low confidence values (those may not be real objects), such predictions with low object confidence values might have contributed noise to the model.  On the other hand, the other features (Faster-COCO, Faster-VG, and YOLO) correspond to objects with high confidence ($\ge0.7$) and hence the noise from the object detection stage is comparatively less.

\begin{table}
\footnotesize
    \centering
    \begin{tabular}{lcc}
    \toprule
     Features & HowManyQA & \makecell{TallyQA-Complex}\\
     \midrule
    Faster-COCO & 1.568&0.942 \\
    Faster-VG &1.562 &0.937 \\
    F-genome&1.62 & 1.02\\
    YOLOv3 &1.560 &0.94 \\
    CenterNet&1.57&0.949\\
     \bottomrule
    \end{tabular}
    \caption{Counting-VQA: Performance (RMSE values) with features from various object-detectors.  \textbf{Lower} is better.  }
    \label{tab:count-feat}
\end{table}
\textbf{(v) Performance of Semantic-dense co-attention on adjectives in the question:} In practical scenarios, the objects may be described by more than one word, i.e., not only its class labels but also some adjectives describing their appearance (e.g., red car).  Since the Semantic-dense co-attention works at the word level and uses only the object class labels (nouns), we conducted experiments for checking whether the model loses the adjective information in the question.  Specifically, we experimented by providing adjectives along with the object's class labels (i.e., ``red car" instead of only ``car").  We used the object-detector from~\cite{bottomup} for this, which provides an additional adjective along with the class labels for each detected object.  We observed a similar performance (1.59 on HowManyQA and 0.99 on TallyQA-complex) as the original model (using the same object-detector, i.e., 1.62 on HowManyQA and 1.02 on TallyQA-complex, see F-genome in Table~\ref{tab:countqa}).  From this experiment, we infer that the Semantic-dense co-attention in the original model provides a sufficient score to the object's adjectives (e.g., ``red" in ``red car") and does not lead to losing such information from the encoded question.  We ascribe this to the following: The pre-trained word embeddings are created by training with a large text corpus.  Hence, the words that often appear together in the text, such as ``adjectives and nouns" (e.g., ``red car," ``big bus," ``blue sea," etc.), will get a relatively similar vector representation than other words. This helps the ``Semantic-dense co-attention" to give a relatively good score to the adjectives as well.

\textbf{(vi) Performance without pre-trained word embeddings:} 
To analyze the performance without using the pre-trained language models, we experimented with one-hot vector representations of object attributes and question words.  Specifically, we manually created a word vocabulary consists of all the unique words in the dataset and created one-hot vectors correspond to the words.  Note that these vectors do not contain any linguistic information.  We experiment with the following cases on the HowManyQA dataset:

\textit{Object attributes and words are in different spaces:} In this experiment, we project both the object attributes and question words to two different learnable spaces (the dimensionality of both spaces is kept the same).  
The performance reduced from 1.56 to 1.74, as expected, since now both the modalities are in two different spaces.  In this case, even though both object attributes and words are in textual space, the two learned spaces being different creates incongruity.  

 \textit{Object attributes and words are in a common learned space:} In this experiment, we project both object-attributes and question words to a common shared learnable space.  
    The results are better than the previous experiment (1.70 instead of 1.74), but there is a notable reduction compared to the original model (1.56 vs.  1.70).  We believe this is because, with this setting, the modalities are in the same space, but the mapping can only leverage the limited linguistic information from the questions obtained while training.  In the original model, both the modalities are in a common linguistically-rich space, and the model can make use of it. 

\begin{table}[t]
    \centering
    \footnotesize
    \begin{tabular}{lcccc}
    \toprule
         Model&yes/no&number&others&overall\\
    \midrule  
    UpDn~\cite{bottomup}&80.3&42.87&55.8&63.2\\
    UpDn+LAT (Glove)&80.52&43.53&59.14&\textbf{64.96}\\
    UpDn+LAT (BERT)&80.6&44.69&60.15&\textbf{65.09}\\
    \midrule
         MUREL~\cite{murel}&80.75&46.72&58.25&65.20\\
         MUREL+LAT(Glove)&80.81&46.85&61.42&\textbf{66.24}\\
         MUREL+LAT(BERT)&80.92&46.95&62.1&\textbf{66.84}\\
         \midrule
         BAN~\cite{ban}&77.88&44.86&60.84&65.39\\
         BAN+LAT(Glove)& 79.67&48.27&62.16&\textbf{67.45}\\
         BAN+LAT (BERT)&79.97&48.31&62.75&\textbf{68.00}\\
         \bottomrule
    \end{tabular}
    \caption{VQA: Performance (VQA accuracy values on the VQAv2-val set) of LAT on various baseline models.  The language-model used in LAT are shown in brackets.
    }
    \label{tab:vqa}
\end{table}
\begin{table}
\footnotesize
    \centering
    \begin{tabular}{lcccccc}
    \toprule
           Model&B@1&B@4&M&R&C&S\\
    \midrule
         UpDn~\cite{bottomup}&79.8&36.3&27.7&56.9&120.1&21.4\\
    \midrule
    UpDn+LAT&\textbf{80.4}&\textbf{37.7}&\textbf{28.4}&\textbf{58.3}&\textbf{127.1}&\textbf{22.0}\\
    \bottomrule
    \end{tabular}
    \caption{Captioning: Performance of LAT on the UpDn baseline~\cite{bottomup}, on the Karpathy-split test set in MSCOCO captions dataset.  B@n,M,R,C and S represent BLEU@n, METEOR, ROUGE-L,CIDEr, and SPICE, respectively.}
    \label{tab:cap}
\end{table}

\subsection{VQA and Captioning}
\label{VQAandCAP}
\begin{figure}[t]
    \centering
    \footnotesize
 \setlength\tabcolsep{.3pt}
    \begin{tabular}{ccc}
   \includegraphics[width=0.29\columnwidth,height=2cm]{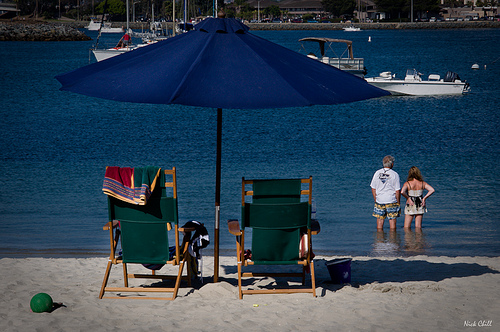}&
    \includegraphics[width=0.29\columnwidth,height=2cm]{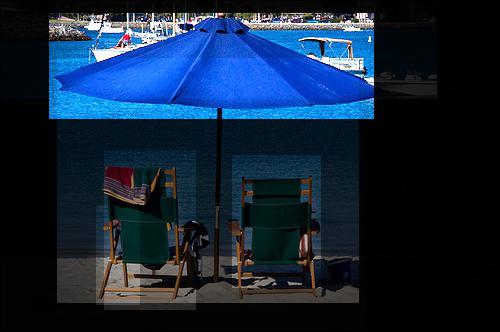}&
    \includegraphics[width=0.29\columnwidth,height=2cm]{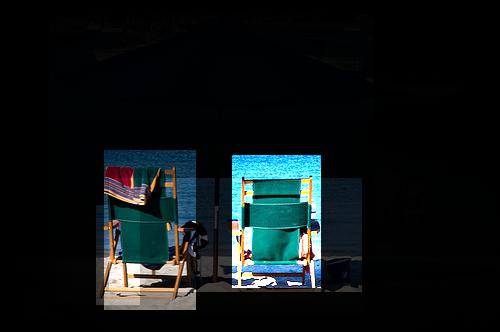} \\
    \makecell[c]{What color are the\\ beach chairs.   \textbf{GT:} green}& \makecell[c]{\textbf{UpDn:} blue} & \makecell[c]{\textbf{+LAT:} green} \\
    \includegraphics[width=0.29\columnwidth,height=2cm]{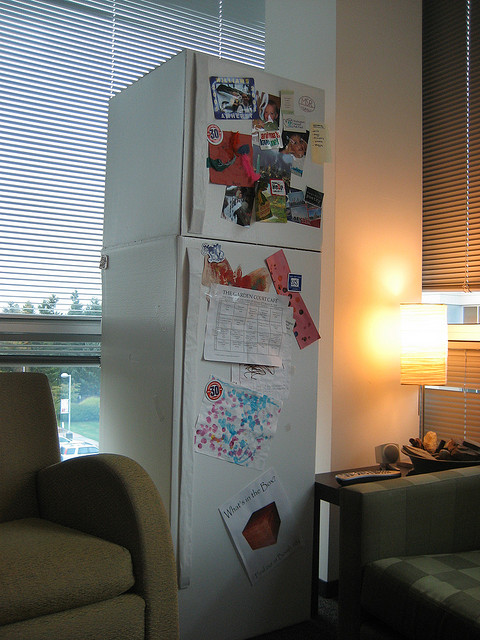}&\includegraphics[width=0.29\columnwidth,height=2cm]{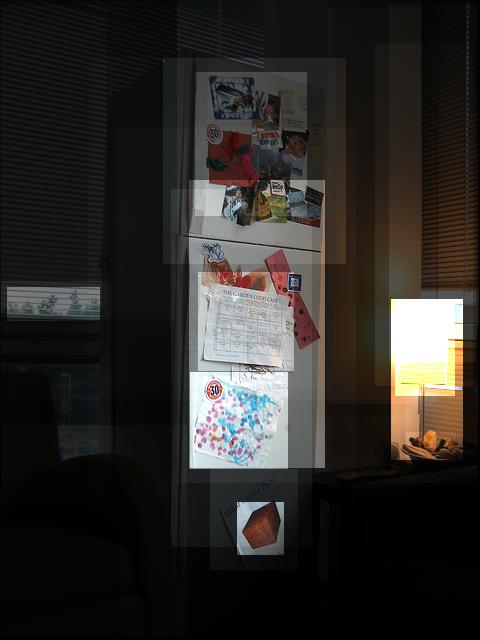}&\includegraphics[width=0.29\columnwidth,height=2cm]{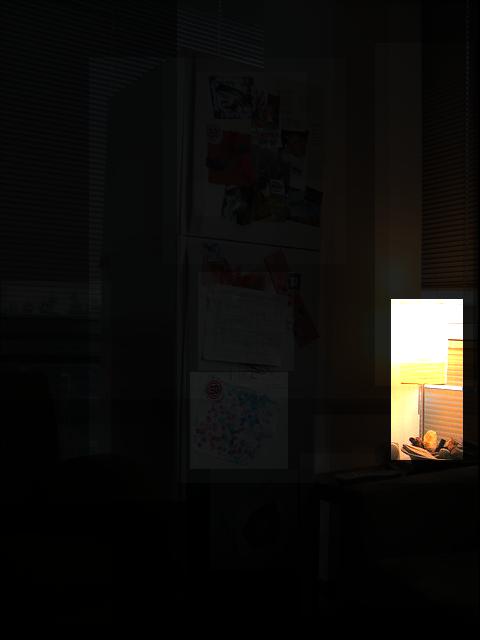} \\
  \makecell[c]{Are the lights on.  \textbf{GT: } yes}&\makecell[c]{\textbf{UpDn:} no}&\makecell[c]{\textbf{+LAT:} yes} \\ 
  \multicolumn{3}{c}{(a) VQA}\\  \includegraphics[width=0.29\columnwidth,height=2cm]{images/mobile_cap.jpg}&\includegraphics[width=0.29\columnwidth,height=2cm]{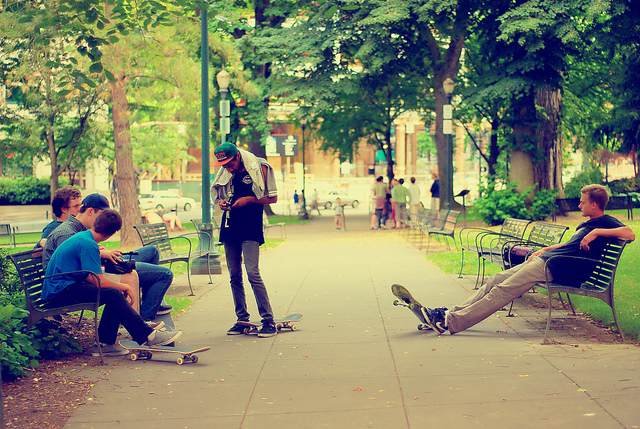} &
     \includegraphics[width=0.29\columnwidth,height=2cm]{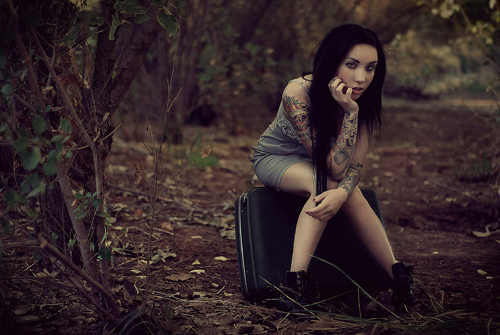} \\
     \makecell[l]{\textbf{UpDn: }a piece of luggage \\sitting on top of a table \\
 \textbf{+LAT: }a close up of \\a cell phone holder
}&\makecell[l]{\textbf{UpDn: }a group of people \\sitting on a park bench \\ \textbf{+LAT:} a group of \\ people sitting on park\\  benches in a park}& \makecell[l]{\textbf{UpDn: } a woman sitting\\  on the ground with  a cell\\ phone \\
 \textbf{+LAT:} a woman sitting on \\ a suitcase in the woods
}\\ 
\multicolumn{3}{c}{(b) Captioning}
      \end{tabular}
    \caption{VQA and Captioning: Qualitative comparisons between UpDn and UpDn+LAT.  }
    \label{fig:updn_qual}
\end{figure}

\textbf{VQA:}
 The quantitative result comparison of the baselines and the same with LAT are shown in Table~\ref{tab:vqa}.  We can see that the incorporation of LAT consistently improves all the models.
 One noteworthy point is the significant margin of improvement achieved in the ``others" category in all the models.  Since the ``others" category has a much wider range of answer distribution (e.g., colors, animals, fruits, vehicles, etc.) than the other two splits (yes/no and number), it requires a more fine-grained and linguistically-aware text-to-image mapping.  The impact of LAT is clearly significant in such cases.  A qualitative result comparison with the UpDn and UpDn+LAT is shown in Fig~\ref{fig:updn_qual} (a).
  We can see that LAT helps the model to focus on the relevant objects more precisely than the baseline.

\textbf{Captioning:}
The quantitative comparison of the UpDn and UpDn+LAT models are shown in Table~\ref{tab:cap}.  We can see that LAT shows improved results over the baseline in all the metrics.  The performance gains in the CIDEr (120.1 to 127.1) and SPICE (21.4 to 22.0) metric show that captions with the LAT are semantically closer to the human-annotated captions than the baseline.  
Some examples depicting LAT's effectiveness in making more image-specific and linguistically robust captions than the baseline are shown in Fig~\ref{fig:updn_qual}(b).  

\section{Conclusion}
In this paper, we proposed a simple yet empirically powerful \textit{``linguistically-aware attention (LAT)"} for reducing the semantic-gap in Vision-language (V-L) tasks.  The notion is to bring rich linguistic-awareness in the attention process by representing both image and text in a common linguistically-rich space.  To facilitate the same, we utilize the object-class labels extracted from the image and words from the text to bring both the modalities to a textual space, then linguistically-aware features are extracted using pre-trained word-embedding networks.  We demonstrated the LAT's impact on three V-L tasks: Counting-VQA, VQA, and image captioning.  In Counting-VQA, we proposed a novel counting-specific VQA model consisting of a semantic-level co-attention mechanism and a count predictor that uses multi-linear algebra to predict the count with a lesser number of parameters.  Our counting model achieved state-of-the-art results on five counting-specific VQA datasets.  In VQA and image captioning, we applied LAT into various baseline models and significantly improved the results in all of them.  We observed that the  LAT improves the semantic robustness of the baseline models.  The LAT is generic and can be incorporated into any other baseline model as well.
 
   At a more generic level, we have explored the transfer learning possibilities from the object-detection and language-modeling research to improve the performance of various tasks in the V-L field.  Based on our study, we conclude that the object class labels obtained from the general-purpose object detectors along with the language models can serve as a semantic bridge between the visual and textual modalities in the V-L tasks.  Such class label features are easy to extract and can be used as a general-purpose standard feature along with the object-level CNN feature for other V-L tasks as well.  While this suggests several directions for future research, an immediate benefit may be obtained in tasks that require fine-grained 
   text-image mapping, such as Visual grounding, Text-based image retrieval, and Semantic segmentation from natural language expressions.  Further, the idea of representing the modalities in a common linguistically-rich space might apply to video-based V-L tasks as well, such as Video-captioning, Video-question answering, and Action localization with language query.  The ideas may find use in correlating other modalities with others as well, such as Speech, IR, Sonar, Radar etc.

\small
\bibliography{mybibfile}
\end{document}